\documentclass{bmcart}

\usepackage{amsthm,amsmath}
\RequirePackage{hyperref}
\usepackage[utf8]{inputenc} 
\usepackage{booktabs}
\usepackage{multirow}
\usepackage{colortbl}
\usepackage{graphicx}
\usepackage{subfig}

\usepackage{amsmath,amssymb,amsfonts}

\usepackage{ulem}
\usepackage{cancel}

\startlocaldefs
\endlocaldefs

\begin{document}

\begin{frontmatter}

\begin{fmbox}
\dochead{Research}


\title{AKI-BERT: a Pre-trained Clinical Language Model for Early Prediction of Acute Kidney Injury}


\author[
   addressref={aff1},
   email={chengsheng.mao@northwestern.edu}
]{\inits{CM}\fnm{Chengsheng Mao}}
\author[
  addressref={aff1},
  email={liang.yao@northwestern.edu}
]{\inits{LY}\fnm{Liang Yao}}
\author[
   addressref={aff1},
  corref={aff1},                       
   email={yuan.luo@northwestern.edu}
]{\inits{YL}\fnm{Yuan Luo} }
\address[id=aff1]{%
  \orgname{Department of Preventive Medicine, Feinberg School of Medicine, Northwestern University},
  \street{750 N Lakeshore Dr.},
  \postcode{60611}
  \city{Chicago},
  \cny{USA}
}
\address[id=aff2]{
  \orgname{Department of Preventive Medicine, Feinberg School of Medicine, Northwestern University},
  \street{750 N Lakeshore Dr.},
  \postcode{60611}
  \city{Chicago},
  \cny{USA}                               
}

\end{fmbox}

\begin{abstractbox}
\begin{abstract} 
Acute kidney injury (AKI) is a common clinical syndrome characterized by a sudden episode of kidney failure or kidney damage within a few hours or a few days. Accurate early prediction of AKI for patients in ICU who are more likely than others to have AKI can enable timely interventions, and reduce the complications of AKI. Much of the clinical information relevant to AKI is captured in clinical notes that are largely unstructured text and requires advanced natural language processing (NLP) for useful information extraction. On the other hand, pre-trained contextual language models such as Bidirectional Encoder Representations from Transformers (BERT) have improved performances for many NLP tasks in general domain recently. However, few have explored BERT on {disease-specific} medical domain tasks such as AKI early prediction. In this paper, we {try to apply BERT to specific diseases and} present an AKI domain-specific pre-trained language model based on BERT (AKI-BERT) that could be used to mine the clinical notes for early prediction of AKI. AKI-BERT is a BERT model pre-trained on the clinical notes of patients having risks for AKI. Our experiments on Medical Information Mart for Intensive Care III (MIMIC-III) dataset demonstrate that AKI-BERT can yield performance improvements for early AKI prediction, thus expanding the utility of the BERT model from general {clinical} domain to {disease-specific} domain.
\end{abstract}


\begin{keyword}
\kwd{acute kidney injury}
\kwd{pre-trained language model}
\kwd{BERT}
\kwd{natural language processing}
\kwd{clinical decision}
\end{keyword}


\end{abstractbox}
%

\end{frontmatter}



\section{Introduction}
Acute kidney injury (AKI) refers to an abrupt decrease in kidney function, resulting in a build-up of waste products in the blood, making it hard to keep the right balance of fluid in the body. AKI can also affect other organs such as the brain, heart, and lungs, thus it is a serious condition that requires immediate treatment. Most of the time, AKI happens in people who are already sick and in the hospital, people who are in the intensive care unit (ICU) are even more likely to develop AKI \cite{ali2007incidence}. AKI usually leads to a prolonged hospital stay and with subsequent morbidity or early mortality post discharge \cite{kellum2013diagnosis,lameire2013contrast,thakar2009incidence,wang2012acute}. Currently, the main biomarker to identify AKI is serum creatinine (SCr) which is a late marker of injury \cite{de2012biomarkers}. However, the efficacy of treatments largely depends on the early prediction of AKI, thus, it is critical to find a way to predict AKI early on to enable timely interventions and treatments.

While structured data in the EHR have obvious value, unstructured clinical notes could contain a wealth of insights into patients like family history, social relations, physician’s comments and many other details that are not likely to be available in the structured records but could suggest onset and progression of a certain disease like AKI. A recent study \cite{hernandez2019real} suggests that real-world data captured in unstructured notes offer more accuracy than structured data in predicting coronary artery disease when trained algorithms are used to mine it. Thus, mining the clinical notes can be an effective way to achieve early prediction of AKI. However, it is challenging to extract predictive information from unstructured text. 

\begin{figure}[!t]
\subfloat[]{\includegraphics[width=.4\textwidth]{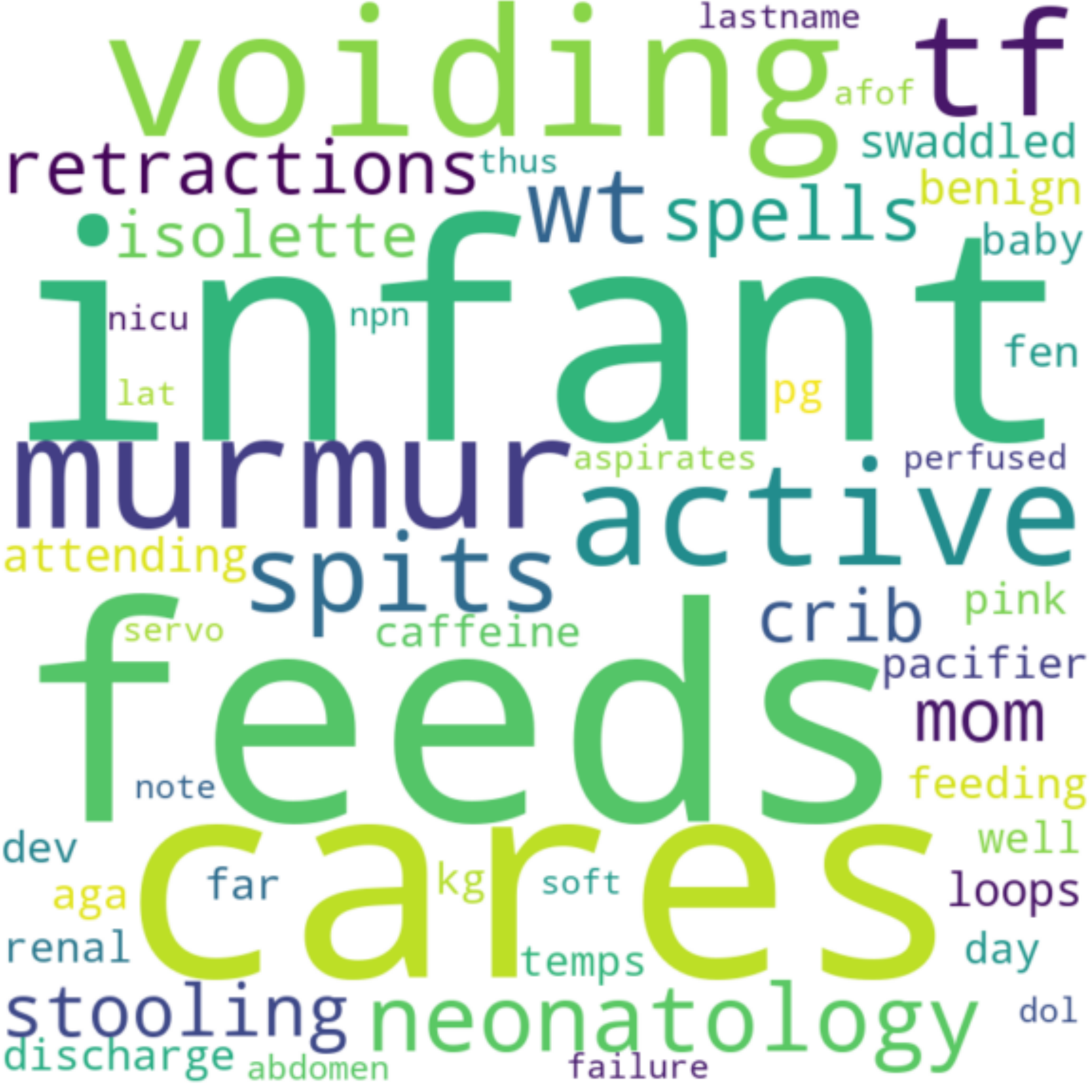} \label{fig:wordcloud_clc}}
\qquad \qquad
\subfloat[]{\includegraphics[width=.4\textwidth]{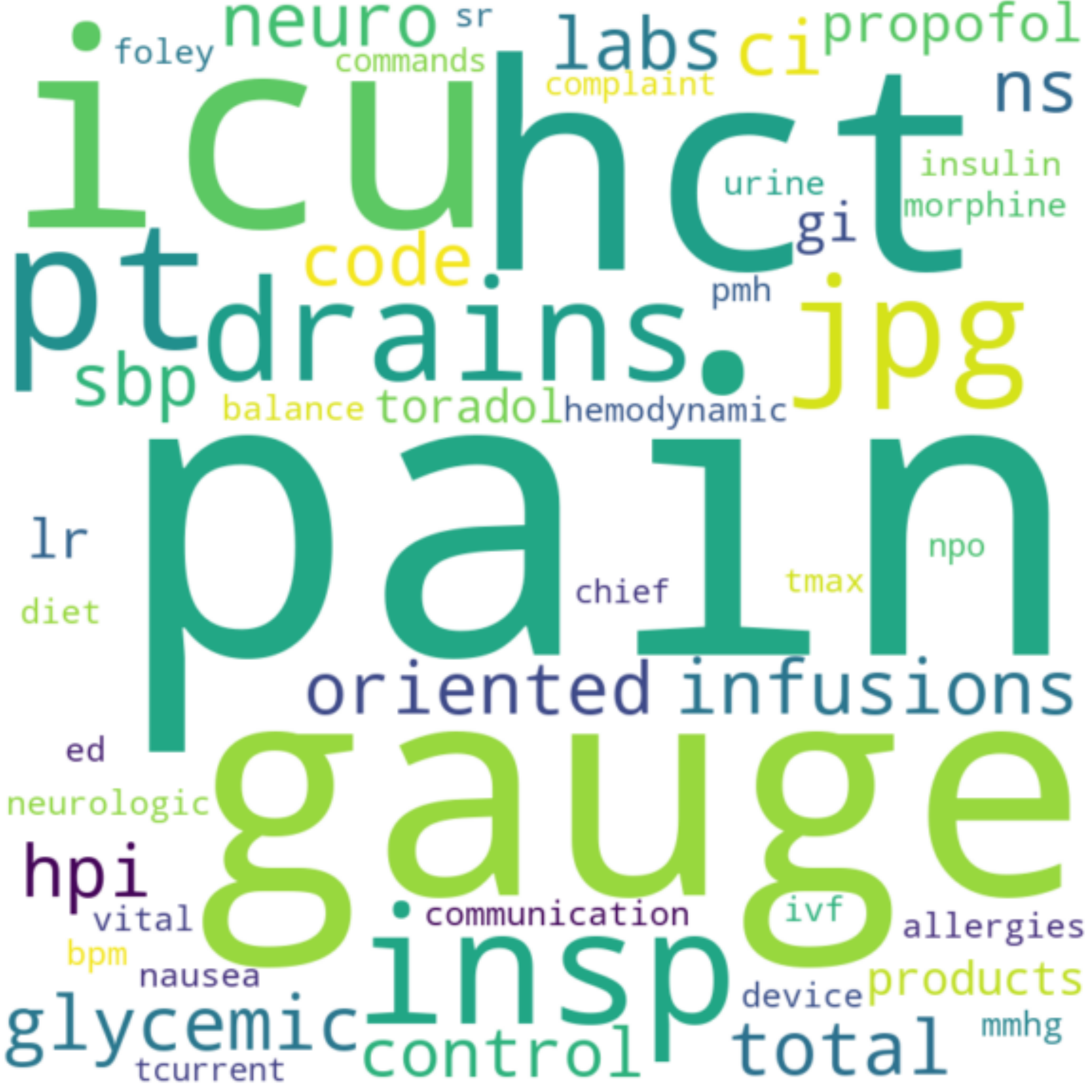} \label{fig:wordcloud_aki}}
\caption{{Word clouds. (a) Words negatively correlated to AKI-related notes; (b)words positively correlated to AKI-related notes.  }}
\label{fig:wordcloud}
\end{figure}

Recently, contextual word embedding models for text mining, such as BERT \cite{devlin2019bert}, ELMo \cite{peters2018deep} and ULMFiT \cite{howard2018universal}, have achieved dramatic successes enabled by transfer learning in many natural language processing (NLP) tasks. However, these models have been primarily explored for general domain text. Recently, BioBERT \cite{lee2019biobert} and clinical BERT \cite{alsentzer2019publicly} are proposed for biomedical text and clinical narratives, respectively. However, clinical notes related to AKI could have some differences from general {clinical} text  in linguistic characteristics, {even if the language difference may not be summarized in a general sense by eyes. We applied a simple logistic regression model to distinguish AKI-related clinical notes and general clinical notes; it can achieve an accuracy of 95\%, suggesting that the AKI related notes could have certain linguistic differences from the general clinical notes, motivating the need for disease domain-specific BERT models for AKI. We also plot the word cloud that are positively and negatively correlated to AKI-related notes by Pearson correlation coefficients in Figure \ref{fig:wordcloud}. } 

Since patients in ICU are most likely to progress to AKI, a comprehensive study of ICU patients is essential to predict the development of AKI. In this paper, we focus on the notes of ICU patients within the first 24 hours of admission for AKI prediction.

\begin{figure}[!t]
\subfloat[Pre-train BERT to AKI-BERT]{\includegraphics[width=.9\textwidth,page=1]{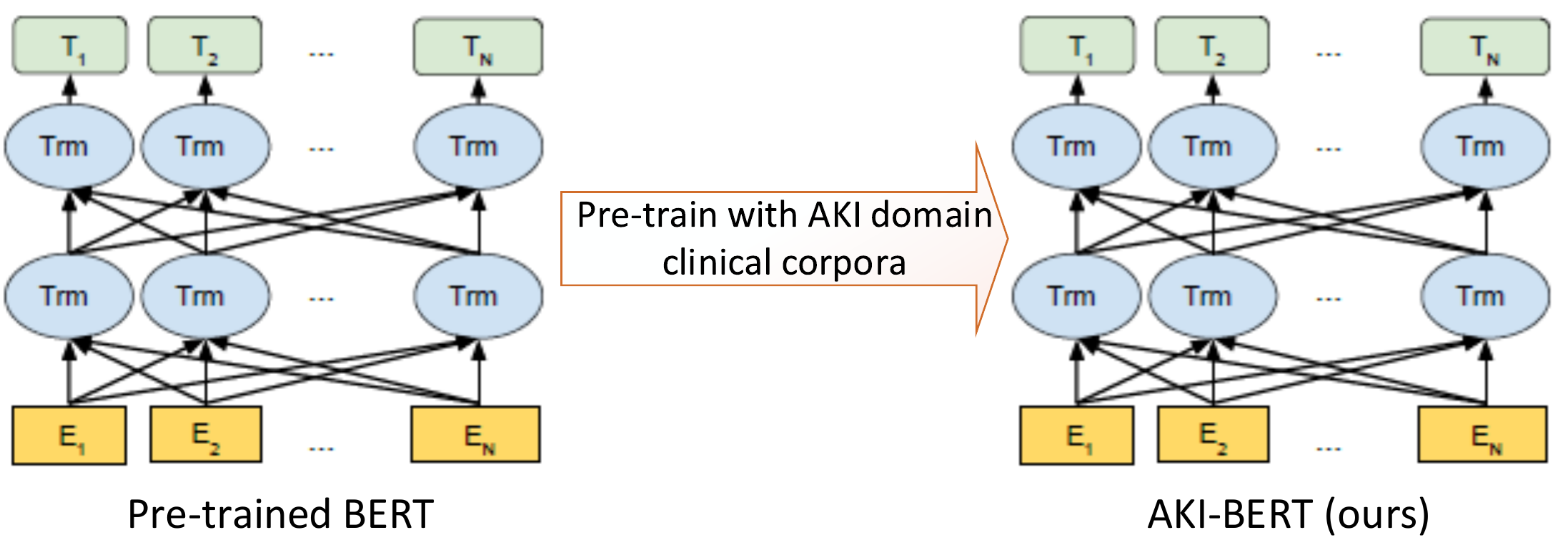} \label{fig:aki-bert}}

\centering
\subfloat[Fine-tune AKI-BERT for AKI prediction]{\includegraphics[width=.9\textwidth,page=2]{flowchat.pdf}\label{fig:akiprediction}}
\caption{Overview of our work. (a) Pre-train BERT to AKI-BERT based on a publicly-available pre-trained model. (b) Fine-tune AKI-BERT for AKI early prediction. AKI-bert is fine-tuned by training the whole model with training notes and the corresponding labels. A test note is input to the model to get the probability of AKI.  }
\label{fig:overview}
\end{figure}

Our methods are illustrated in Fig. \ref{fig:overview}. In our methods, three steps are taken to achieve the risk prediction of AKI, (1) we further pre-train the publicly-released pre-trained BERT model (e.g., BERT-base \cite{devlin2019bert}, BioBert \cite{lee2019biobert}, Bio+Clinical BERT \cite{alsentzer2019publicly}) with unlabeled AKI domain notes, and get AKI-BERT (Fig. \ref{fig:aki-bert}); (2) we fine-tune AKI-BERT for an AKI early prediction task with the training notes and the corresponding labels (Fig. \ref{fig:akiprediction}); (3) the fine-tuned AKI-BERT model is used to predict the probability that a patient will develop AKI from the notes of the patient (Fig. \ref{fig:akiprediction}). 

The main contribution of this study are summarized as follows:

\begin{itemize}
\item We present AKI-BERT for AKI-related clinical NLP tasks. To the best of our knowledge, {this is the first study to pretrain and release a contextual language model in a disease-specific domain}, i.e., AKI. We release 3 varieties of AKI-BERT models that were obtained by pre-training BERT-base \cite{devlin2019bert}, BioBERT \cite{lee2019biobert}, and Bio+Clinical BERT (BC-BERT) \cite{alsentzer2019publicly} on AKI-related notes, respectively. The code and pre-trained models can be available at \url{https://github.com/mocherson/AKI_bert}.
\item We demonstrate that AKI-BERT models pre-trained on AKI-related notes improve the performance of AKI prediction compared to BERT models pre-trained on general domain and clinical domains.
\end{itemize}

\section{Related Work}
\subsection{AKI Prediction}
The comprehensive and large amounts of electronic health records (EHR) data enables AKI prediction by developing data-driven models \cite{wang2012acute}. Most of the previous studies focused on specific patient populations such as cardiac surgery patients \cite{zhou2016development,o2016acute,kim2013simplified}, elderly patients \cite{kate2016prediction} and pediatrics patients \cite{sanchez2016development}. Some other studies of AKI prediction focused on the validation of novel biomarkers or risk scores \cite{perazella2015urine,kerr2014developing, zhou2016development}. Most of these studies did not exclude the patients that have already progressed to AKI, where therapeutic interventions usually have limited effectiveness \cite{colpaert2012impact}. Our approach focuses on patients who do not have AKI when admit to ICU to achieve the early prediction of AKI. Recently, Li et al. \cite{li2018early} tried to early predict AKI with clinical notes using bag-of-words features. Toma{\v{s}}ev et al. \cite{tomavsev2019clinically} provided a notable example of using lab values for continuous prediction of AKI. Sun et al. \cite{sun2019early} combined structured EHR and unstructured clinical notes for early prediction of AKI. Different from previous research, this paper presents a pre-trained language model on AKI-related domain, i.e., AKI-BERT. The pre-trained model can be fine-tuned for AKI early prediction. This paper primarily focuses on the task of early AKI prediction, and leave other NLP tasks for future work.

\subsection{Text Classification}
Predicting AKI for a clinical note can be seen as a text classification task where a classifier is trained to accept a note text as input and outputs the risk the patient of the note will progress to AKI. Traditional text classification studies mainly focus on extracting features from the texts and training a classifier for the extracted features. Many kinds of features can be extracted from text, such as the popular bag-of-words features, n-grams~\cite{wang2012baselines} and entities in ontologies~\cite{chenthamarakshan2011concept}. Some other studies tried to convert texts to graphs and perform feature engineering on graphs and subgraphs~\cite{rousseau2015text,skianis2016regularizing,luo2014automatic,luo2015subgraph}. 

Modern text classification usually applies deep neural networks to automatically learn a vector representation for a text for classification. Many kinds of neural networks were explored for text classification, such as convolutional neural networks (CNN) \cite{kim2014convolutional,yao2019clinical,zhang2015character,conneau2017very}, recurrent neural networks (RNN) \cite{tai2015improved,liu2016recurrent,luo2017recurrent} and graph convolutional networks (GCN) \cite{yao2019graph}. To further increase the representation flexibility of deep learning models, attention mechanisms have been introduced as a part of the models for text classification \cite{yang2016hierarchical,wang2016attention,vaswani2017attention,velickovic2018graph}.

\subsection{Contextual Word Embedding Models}
Recent studies showed that the success of deep learning for text classification largely depends on the effectiveness of the word embeddings, i.e., the word-level representations \cite{Shen2018Baseline,joulin2017bag,wang2018joint}. Traditional
word-level vector representations, such as word2vec \cite{mikolov2013distributed}, GloVe \cite{pennington2014glove}, and fastText \cite{bojanowski2017enriching}, represent a word as a single vector and cannot differentiate the different word senses of the same word based on the surrounding context. Recently, contextual word embedding models such as ELMo \cite{peters2018deep} and BERT \cite{devlin2019bert} showed their superiority in many NLP tasks by providing contextualized word representations. These models can create a context-sensitive embedding for each word in a given sentence by pre-training on a large text corpus. BERT has been found to be even more effective than ELMo on a variety of tasks, including those in the clinical domain \cite{si2019enhancing,yao2019kg}. This is because the deeper architecture and more parameters could generate more powerful representations. For this reason, we only consider the BERT architecture in this paper. 

There are several studies exploring the utility of BERT in the clinical or biomedical domains. BioBERT \cite{lee2019biobert} pre-trained the BERT model on a biomedical domain corpus sourced from PubMed article abstracts and PubMed Central article full texts. They found BioBERT could significantly outperform BERT on three representative biomedical text mining tasks including named entity recognition, relation extraction and question answering. On clinical text, Si et al. \cite{si2019enhancing} trained a BERT language model on clinical note corpora and yielded improvements over both traditional word embeddings and ELMo embeddings on four datasets for concept extraction. Alsentzer et al. \cite{alsentzer2019publicly} pre-trained BERT and BioBERT on clinical text for a clinical domain-specific BERT model, and yielded performance improvements on three common clinical NLP tasks as compared to non-domain-specific models. Huang et al. also pre-trained BERT on MIMIC-III clinical notes and fine-tuned BERT for readmission prediction \cite{huang2019clinicalbert}. Yao et al. \cite{yao2019traditional} pre-trained BERT in Chinese on a corpus of traditional Chinese medicine (TCM) clinical records, and achieved the state-of-the-art performance for TCM clinical records classification. {Laila et al. \cite{rasmy2021med} adapted BERT to the structured EHR domain and proposed Med-BERT for disease prediction; they evaluated Med-BERT for heart failure and pancreatic cancer and achieved substantial improvements.} However, no BERT models have been {pre-trained specifically} for AKI domain in literature. This paper will pre-train a BERT model on AKI-related domain, and apply the model to early prediction of AKI. 

\section{Methods}
\subsection{BERT: Bidirectional Encoder Representations from Transformers}
BERT \cite{devlin2019bert} is a contextual word representation model that is developed on a multi-layer bidirectional transformer encoder which is based on a self-attention mechanism \cite{vaswani2017attention}. BERT is pre-trained with a masked language model (MLM) that predicts randomly masked words in a sequence, and hence can be used for learning better bidirectional representations than unidirectional language models (i.e., left-to-right and right-to-left). BERT is also pre-trained for a binary next sentence prediction (NSP) task to model the relationship between two sentences. By pre-training the model for the two tasks on a large text corpus, BERT obtains state-of-the-art performance on most NLP tasks with minimal task-specific architecture modifications.

BERT-base is obtained by pre-training the original 12-layer BERT model  on general domain text corpora, English Wikipedia and BooksCorpus \cite{devlin2019bert}. BioBERT \cite{lee2019biobert} is obtained by pre-training BERT-base on large-scale biomedical corpora sourced from PubMed article abstracts and PubMed Central article full texts. Clinical BERT and Bio+Clinical BioBERT (BC-BERT) are obtained by respectively pre-training BERT-base and BioBERT on approximately 2 million clinical notes from the MIMIC-III v1.4 database. In this paper, we further pre-train previously pre-trained BERT models, including BERT-base, BioBERT and BC-BERT, on AKI-related notes from MIMIC-III to obtain AKI-BERT. ALL these models have the same structure and are all optimized by minimizing the loss of the two tasks, i.e., MLM and NSP. The differences are parameters that are learned with different parameter initialization and training data. 

\subsection{AKI-BERT: AKI domain pre-training}
The AKI-related notes used to pre-train AKI-BERT is extracted from MIMIC-III dataset. According to the guidelines of Kidney Disease Improving Global Outcomes (KDIGO) \cite{kidney2009kdigo}, AKI is defined by a certain increase of serum creatinine (SCr), thus, {Following previous work \cite{li2018early},} we consider the patients who had a creatinine measured in 72 hours following ICU admission. {In order to avoid data leakage,} Patients with history of {AKI} or chronic kidney disease (CKD) are excluded, {patients whose clinical notes have mentioned kidney dysfunction related words and their abbreviations are also excluded}. The notes during the first 24 hours of ICU admission for the {remaining} patients are used as the AKI corpus to pre-train AKI-BERT. Finally, The AKI corpus contains 16,560 notes, 2,838,906 sentences and 49,738,920 tokens tokenized by WordPiece \cite{wu2016google}. 

AKI domain pre-training for AKI-BERT is actually training the model with parameters initialized with a previously pre-trained BERT model on AKI corpus. The losses of the two tasks MLM and NSP are summed to optimize the model on the training corpus \cite{devlin2019bert}.

\begin{figure}[!t]


{\includegraphics[width=.55\linewidth,page=3]{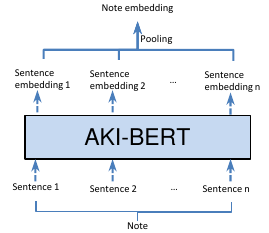} \label{fig:pooling}}
\caption{{Task-specific fine-tuning with Pooling strategies.}}
\label{fig:finetune}
\end{figure}

\subsection{AKI Prediction: Task-specific Fine-tuning}
For a specific downstream task, after the pre-training of BERT, a task-specific fine-tuning step should be applied, where BERT is initialized with the pre-trained parameters, new layers can be added to the basic architecture for the specific task, all of the parameters are updated by minimizing the task-specific loss.

\subsubsection{Data}
Based on the criteria of KDIGO, {following previous work \cite{zimmerman2019early},} a patient ICU stay is {labeled} as AKI if either of the following two conditions is met: 1) increase in serum creatinine is greater than or equal to 0.3 mg/dL (i.e., 26.5 micromol/L) within 48 hours; 2) increase in serum creatinine is greater than or equal to 1.5 times baseline. {2531 ICU stays met condition 1 and 1847 ICU stays met condition 2, with 1593 overlaps.} The baseline serum creatinine is defined by the minimum creatinine value in the first ICU day. {All other ICU stays that had a creatinine measurement are labeled as non-AKI. Finally, we have a total of 2785 AKI ICU stays and 13775 non-AKI ICU stays.} All the {labeled} ICU stays with the corresponding notes are randomly split into training set, validation set and test set with notes of the same ICU stay in the same set. The notes in training set together with the corresponding labels (i.e., whether AKI occurs) are used to fine-tune the model end-to-end. The validation set is used to select the best model for evaluation in the training process. The test set is used to evaluate the fine-tuned model. The count information of notes in the three splits is shown in Table \ref{tab:dataset}.

\begin{table}[!t]
  \centering
  \caption{Note count information in training, validation, and test set.}
    \begin{tabular}{lcccc}
    \toprule
          & All & AKI & Non-AKI & AKI prevalance \\
    \midrule
    training & 9248  & 1542  & 7706 & 16.67\% \\
    validation & 2312  & 385   & 1927 & 16.65\% \\
    test  & 5000  & 858   & 4142 & 17.16\% \\
    \midrule
    overall & 16560 & 2785  & 13775 & 16.82\% \\
    \bottomrule
    \end{tabular}%
  \label{tab:dataset}%
\end{table}%

As we can see from Table \ref{tab:dataset}, the dataset is imbalanced, only a few of the samples ($\sim17\%$) correspond to AKI onset. Training on an imbalanced dataset will bias the classifier to the majority class. Even if a great weight is given to the minority class in the loss function, the issue still exists. Because a batch randomly selected from the data usually contains no minority samples, especially in a small batch. We have tried three schemes to address this issue independently in our experiments. (1) Stratified Batch Sampler (SBS): we develop a stratified batch sampler algorithm to ensure each batch contains an equal proportion of samples from each class, and use a class-weighted loss function. (2) Down Sampling (DS): we downsample the majority class to make it balanced with the minority class by random choice without replacement. (3) Up Sampling (US): we upsample the minority class to make it balanced with the majority class by random choice with replacement.

\subsubsection{Classifier fine-tuning}
Classifier fine-tuning for AKI prediction is to train a classifier constructed with the BERT structure and a linear layer used for classification. {The BERT can accept a single sentence as input and output the sentence embedding.} For a general single sentence classification task, the sentence embedding is fed to a linear layer to output the class probabilities which are combined with the true class label to compute the loss for optimization. 

Since BERT can accept a maximum sequence length of 512, and a clinical note usually has much more words. A long note cannot be directly handled by BERT. We adopt two strategies to address this issue. (1) Truncating: each note is considered as a sequence, long notes are truncated to 512 tokens. (2) Pooling: we split each note into sentences that do not exceed the maximum sequence length, and get an embedding for each sentence by inputting the sentences to AKI-BERT. We use a pooling approach to get the embedding of the note based on the sentence embeddings (Fig. \ref{fig:finetune}). In our experiments, we use MaxPooling where the sentence embeddings are aggregated to one note embedding by retaining the maximum value of each dimension of the sentence embeddings. {We acknowledge that there may be a lot of even better polling strategies or models (e.g., Longformer \cite{beltagy2020longformer}) for long notes embedding. In this study, we focused on BERT model for a specific disease, the pooling strategies or models for long texts are not the focus of this paper. We will leave the discussion of Longformer and pooling strategy for long text for the future work. } 

Note that the AKI corpus for pre-training contains all the clinical text without any label information. The fine-tuning uses the training set that is a part of the AKI corpus together with the label information to train the model for the AKI early prediction task. The evaluation of the model is based on the test set and the corresponding labels. Since in the model pre-training process, the test notes are seen but the label of the test notes are unknown, the whole learning process is in the semi-supervised learning framework which presents a valid evaluation of the model.

\section{Experiments}
\subsection{Settings}
For AKI-domain pre-training, we use the default settings as BERT implemented in pytorch transformers \cite{pytorch_tranformer} \footnote{\url{https://github.com/huggingface/transformers/blob/v1.0.0/examples/lm_finetuning/simple_lm_finetuning.py}}, except for the initialization and corpus. We obtain 3 varieties of BERT models on AKI-related corpus by pre-training BC-BERT, BioBERT and BERT-base, named AKI-BC-BERT, AKI-BioBERT and AKI-baseBERT, respectively.

For task-specific fine-tuning, the truncating setting is actually the same as the original BERT. For pooling setting, we set max sequence length as 32, since the average sentence length is 17.6 in our corpus. Some notes have too many sentences for our GPU memory to process, we set a note can have 180 sentences at most, notes with more than 180 sentences are randomly sampled to 180 sentences. Our batch size is set as 4 and the training epoch is 5. We evaluate the model with validation set every 500 batches in the training process. The model with maximum AUC in validation set is selected for AKI prediction on the test set.

\subsection{Results and Discussions}
The AKI prediction performances of the models on the test set are listed in Table \ref{tab:results}. {The setting weight just applies a class weight that is inversely proportional to the sample size in the class.} Static setting does not do classifier fine-tuning and just use the output note embedding as features fed to a logistic regression classifier, {this can be considered only the linear classifier is trained during the fine-tuning phase. All other settings, including SBS, DS, US and weight, allowed all parameters from BERT and the linear layer to be fine-tuned in the classifier fine-tuning process.}  Since our goal is to demonstrate that AKI domain-specific embeddings can be more effective than general embeddings, thus some methods that are not based on word embeddings (e.g., bag-of-words) are not compared with in our experiments. From Table \ref{tab:results}, overall, AKI-BERT variants perform better than others for each setting in terms of AUC and F1, the reason is that the pre-training of AKI-BERT incorporates general information in AKI domain into AKI-BERT, thus can achieve more precise representations for AKI domain words and more effective for AKI domain tasks. 

In Table \ref{tab:results}, we note some models fail in fine-tuning for AKI prediction, {NPV=nan means all samples are assigned to AKI class, where Recall=1 and Precision=0.172 (858/5000). Precision=nan means all samples are assigned to non-AKI class, where Recall=0 and SPC=1.} Since all the BERT models have the same architecture, and the only differences are the initialization which affects the optimization of the models. The models that are obtained by pre-training on the AKI corpus are expected to have more precise initialization including the initial word embeddings for AKI-related tasks. A bad initialization may cause the optimization of the model stuck on local optimum or plateaus, thus get a bad performance. This is why model pre-training is important.

\begin{table}[!t]
\tiny
  \centering
  \caption{{Performance of different BERT models for AKI prediction. SBS: stratified batch sampling; DS: down sampling; US: up sampling; SPC: specificity; PPV: positive predictive value; NPV: negative predictive value. Best AUC and F1 for each setting are bolded. 95\% confidence interval is in the bracket.}}
    \begin{tabular}{l|l|llllll}
    \toprule
    \textbf{Setting} & \multicolumn{1}{c|}{\textbf{Model}} & \multicolumn{1}{c}{\textbf{AUC}} & \multicolumn{1}{c}{\textbf{Precision/PPV}} & \multicolumn{1}{c}{\textbf{Recall/Sensitivity}} & \multicolumn{1}{c}{\textbf{F1}} & \multicolumn{1}{c}{\textbf{SPC}} & \multicolumn{1}{c}{\textbf{NPV}} \\
    \midrule
    \multirow{7}[0]{*}{DS+Pooling} & AKI-baseBERT & 0.747 (0.728-0.766) & 0.356 (0.332-0.380) & 0.619 (0.586-0.651) & 0.452 (0.427-0.477) & 0.768 (0.755-0.781) & 0.907 (0.897-0.916) \\
          & AKI-BC-BERT & 0.757 (0.739-0.775) & 0.353 (0.329-0.377) & 0.648 (0.616-0.680) & 0.457 (0.432-0.482) & 0.754 (0.741-0.768) & 0.912 (0.902-0.921) \\
          & AKI-BioBERT & \textbf{0.761} (0.742-0.779) & 0.370 (0.345-0.395) & 0.624 (0.590-0.655) & \textbf{0.465} (0.439-0.490) & 0.780 (0.767-0.793) & 0.909 (0.900-0.918) \\
          & BC-BERT & 0.758 (0.740-0.777) & 0.366 (0.342-0.390) & 0.639 (0.606-0.671) & \textbf{0.465} (0.440-0.490) & 0.771 (0.758-0.783) & 0.911 (0.902-0.921) \\
          & BERT-base & 0.566 (0.544-0.586) & 0.172 (0.161-0.182) & 1.000 (1.000-1.000) & 0.293 (0.278-0.308) & 0.000 (0.000-0.000) & nan (nan-nan) \\
          & BioBERT & 0.738 (0.719-0.757) & 0.275 (0.257-0.294) & 0.723 (0.693-0.753) & 0.399 (0.377-0.420) & 0.606 (0.591-0.620) & 0.913 (0.903-0.924) \\
          & Clinical BERT & 0.748 (0.729-0.767) & 0.328 (0.306-0.350) & 0.683 (0.651-0.714) & 0.443 (0.419-0.466) & 0.710 (0.695-0.724) & 0.915 (0.906-0.925) \\
    \midrule
    \multirow{7}[0]{*}{SBS+Pooling} & AKI-baseBERT & 0.759 (0.741-0.778) & 0.343 (0.321-0.366) & 0.670 (0.639-0.703) & 0.454 (0.430-0.478) & 0.734 (0.720-0.747) & 0.915 (0.905-0.924) \\
          & AKI-BC-BERT & 0.760 (0.742-0.779) & 0.325 (0.304-0.346) & 0.711 (0.680-0.741) & 0.446 (0.423-0.469) & 0.695 (0.680-0.708) & 0.921 (0.911-0.930) \\
          & AKI-BioBERT & \textbf{0.762} (0.743-0.781) & 0.377 (0.353-0.402) & 0.676 (0.644-0.707) & \textbf{0.484} (0.459-0.509) & 0.768 (0.756-0.781) & 0.920 (0.910-0.928) \\
          & BC-BERT & 0.760 (0.741-0.778) & 0.333 (0.311-0.355) & 0.704 (0.673-0.734) & 0.452 (0.429-0.475) & 0.708 (0.694-0.721) & 0.920 (0.911-0.929) \\
          & BERT-base & 0.680 (0.659-0.701) & nan (nan-nan) & 0.000 (0.000-0.000) & 0.000 (0.000-0.000) & 1.000 (1.000-1.000) & 0.828 (0.818-0.839) \\
          & BioBERT & 0.566 (0.545-0.587) & nan (nan-nan) & 0.000 (0.000-0.000) & 0.000 (0.000-0.000) & 1.000 (1.000-1.000) & 0.828 (0.818-0.839) \\
          & Clinical BERT & 0.760 (0.741-0.778) & 0.347 (0.324-0.370) & 0.675 (0.644-0.706) & 0.458 (0.434-0.482) & 0.737 (0.723-0.750) & 0.916 (0.907-0.925) \\
          \midrule
    \multirow{7}[0]{*}{SBS+Truncating} & AKI-baseBERT & 0.738 (0.718-0.757) & 0.430 (0.401-0.459) & 0.552 (0.519-0.586) & \textbf{0.483} (0.456-0.510) & 0.848 (0.837-0.859) & 0.901 (0.892-0.911) \\
          & AKI-BC-BERT & 0.737 (0.718-0.757) & 0.401 (0.374-0.427) & 0.603 (0.569-0.635) & 0.481 (0.454-0.507) & 0.813 (0.802-0.825) & 0.908 (0.899-0.917) \\
          & AKI-BioBERT & \textbf{0.751} (0.731-0.770) & 0.414 (0.386-0.443) & 0.570 (0.537-0.604) & 0.479 (0.452-0.506) & 0.833 (0.821-0.844) & 0.903 (0.894-0.913) \\
          & BC-BERT & 0.708 (0.688-0.727) & 0.000 (0.000-0.000) & 0.000 (0.000-0.000) & 0.000 (0.000-0.000) & 0.999 (0.998-1.000) & 0.828 (0.818-0.839) \\
          & BERT-base & 0.702 (0.682-0.723) & 0.172 (0.161-0.182) & 1.000 (1.000-1.000) & 0.293 (0.278-0.308) & 0.000 (0.000-0.000) & nan (nan-nan) \\
          & BioBERT & 0.513 (0.494-0.532) & nan (nan-nan) & 0.000 (0.000-0.000) & 0.000 (0.000-0.000) & 1.000 (1.000-1.000) & 0.828 (0.818-0.839) \\
          & Clinical BERT & 0.711 (0.690-0.731) & 0.515 (0.457-0.573) & 0.178 (0.153-0.205) & 0.265 (0.231-0.298) & 0.965 (0.959-0.971) & 0.850 (0.840-0.860) \\
          \midrule
    \multirow{7}[0]{*}{US+Pooling} & AKI-baseBERT & 0.746 (0.726-0.765) & 0.373 (0.348-0.399) & 0.614 (0.581-0.647) & \textbf{0.465} (0.438-0.489) & 0.787 (0.774-0.799) & 0.908 (0.898-0.917) \\
          & AKI-BC-BERT & \textbf{0.756} (0.737-0.774) & 0.323 (0.302-0.344) & 0.710 (0.680-0.740) & 0.444 (0.421-0.467) & 0.692 (0.678-0.706) & 0.920 (0.910-0.929) \\
          & AKI-BioBERT & 0.742 (0.723-0.762) & 0.269 (0.252-0.287) & 0.758 (0.728-0.786) & 0.397 (0.376-0.418) & 0.574 (0.559-0.589) & 0.920 (0.909-0.930) \\
          & BC-BERT & 0.738 (0.719-0.758) & 0.285 (0.266-0.304) & 0.725 (0.695-0.755) & 0.409 (0.386-0.431) & 0.624 (0.608-0.638) & 0.916 (0.906-0.926) \\
          & BERT-base & 0.651 (0.630-0.672) & 0.206 (0.192-0.220) & 0.773 (0.745-0.800) & 0.325 (0.307-0.344) & 0.384 (0.369-0.398) & 0.891 (0.876-0.905) \\
          & BioBERT & 0.734 (0.715-0.752) & 0.283 (0.265-0.302) & 0.746 (0.716-0.775) & 0.410 (0.388-0.432) & 0.608 (0.594-0.623) & 0.920 (0.910-0.930) \\
          & Clinical BERT & 0.749 (0.730-0.767) & 0.344 (0.321-0.367) & 0.659 (0.626-0.690) & 0.452 (0.427-0.476) & 0.740 (0.726-0.753) & 0.913 (0.903-0.922) \\
          \midrule
    \multirow{7}[0]{*}{weight+Pooling} & AKI-baseBERT & 0.748 (0.729-0.767) & 0.469 (0.435-0.502) & 0.469 (0.435-0.502) & 0.469 (0.439-0.498) & 0.890 (0.880-0.900) & 0.890 (0.880-0.899) \\
          & AKI-BC-BERT & 0.747 (0.727-0.766) & 0.457 (0.425-0.490) & 0.484 (0.450-0.518) & 0.470 (0.441-0.499) & 0.881 (0.871-0.891) & 0.892 (0.882-0.901) \\
          & AKI-BioBERT & \textbf{0.761} (0.742-0.779) & 0.468 (0.437-0.501) & 0.509 (0.476-0.543) & \textbf{0.488} (0.459-0.516) & 0.880 (0.870-0.890) & 0.896 (0.887-0.906) \\
          & BC-BERT & 0.596 (0.576-0.617) & nan (nan-nan) & 0.000 (0.000-0.000) & 0.000 (0.000-0.000) & 1.000 (1.000-1.000) & 0.828 (0.818-0.839) \\
          & BERT-base & 0.559 (0.538-0.580) & 0.172 (0.161-0.182) & 1.000 (1.000-1.000) & 0.293 (0.278-0.308) & 0.000 (0.000-0.000) & nan (nan-nan) \\
          & BioBERT & 0.537 (0.515-0.558) & 0.172 (0.161-0.182) & 1.000 (1.000-1.000) & 0.293 (0.278-0.308) & 0.000 (0.000-0.000) & nan (nan-nan) \\
          & Clinical BERT & 0.554 (0.533-0.575) & nan (nan-nan) & 0.000 (0.000-0.000) & 0.000 (0.000-0.000) & 1.000 (1.000-1.000) & 0.828 (0.818-0.839) \\
          \midrule
    \multirow{7}[0]{*}{static+Pooling} & AKI-baseBERT & 0.725 (0.705-0.745) & 0.308 (0.287-0.329) & 0.634 (0.601-0.666) & 0.414 (0.391-0.438) & 0.704 (0.690-0.718) & 0.903 (0.893-0.913) \\
          & AKI-BC-BERT & 0.737 (0.717-0.756) & 0.340 (0.318-0.363) & 0.642 (0.610-0.674) & \textbf{0.445} (0.421-0.469) & 0.742 (0.728-0.755) & 0.909 (0.900-0.919) \\
          & AKI-BioBERT & \textbf{0.745} (0.725-0.763) & 0.342 (0.319-0.365) & 0.640 (0.608-0.671) & \textbf{0.445} (0.421-0.469) & 0.745 (0.731-0.758) & 0.909 (0.899-0.918) \\
          & BC-BERT & 0.733 (0.714-0.751) & 0.313 (0.292-0.334) & 0.649 (0.617-0.680) & 0.422 (0.399-0.446) & 0.705 (0.691-0.719) & 0.907 (0.896-0.916) \\
          & BERT-base & 0.692 (0.672-0.712) & 0.291 (0.270-0.312) & 0.617 (0.584-0.650) & 0.395 (0.371-0.419) & 0.689 (0.675-0.703) & 0.897 (0.886-0.907) \\
          & BioBERT & 0.708 (0.688-0.728) & 0.293 (0.272-0.313) & 0.649 (0.617-0.681) & 0.403 (0.380-0.426) & 0.675 (0.660-0.689) & 0.903 (0.892-0.913) \\
          & Clinical BERT & 0.715 (0.695-0.735) & 0.308 (0.287-0.329) & 0.648 (0.615-0.680) & 0.418 (0.394-0.441) & 0.699 (0.685-0.713) & 0.906 (0.895-0.916) \\
          \midrule
    \multirow{7}[0]{*}{static+Truncating} & AKI-baseBERT & 0.736 (0.716-0.755) & 0.332 (0.309-0.354) & 0.650 (0.619-0.682) & 0.439 (0.415-0.463) & 0.728 (0.715-0.742) & 0.910 (0.900-0.919) \\
          & AKI-BC-BERT & 0.751 (0.732-0.769) & 0.344 (0.321-0.367) & 0.679 (0.648-0.710) & \textbf{0.457} (0.432-0.481) & 0.732 (0.718-0.745) & 0.917 (0.907-0.926) \\
          & AKI-BioBERT & \textbf{0.752} (0.733-0.770) & 0.346 (0.323-0.369) & 0.659 (0.626-0.690) & 0.454 (0.429-0.478) & 0.742 (0.729-0.755) & 0.913 (0.903-0.922) \\
          & BC-BERT & 0.733 (0.713-0.753) & 0.313 (0.292-0.334) & 0.672 (0.641-0.704) & 0.427 (0.403-0.450) & 0.694 (0.680-0.708) & 0.911 (0.901-0.921) \\
          & BERT-base & 0.703 (0.683-0.723) & 0.283 (0.263-0.304) & 0.632 (0.600-0.664) & 0.391 (0.368-0.414) & 0.669 (0.654-0.683) & 0.898 (0.887-0.908) \\
          & BioBERT & 0.714 (0.694-0.733) & 0.289 (0.269-0.309) & 0.654 (0.621-0.686) & 0.401 (0.378-0.424) & 0.666 (0.652-0.681) & 0.903 (0.892-0.913) \\
          & Clinical BERT & 0.717 (0.698-0.736) & 0.295 (0.274-0.315) & 0.659 (0.627-0.690) & 0.407 (0.383-0.430) & 0.674 (0.659-0.688) & 0.905 (0.894-0.915) \\
          \bottomrule
    \end{tabular}%
  \label{tab:results}%
\end{table}%

BC-BERT is obtained by pre-training BioBERT on the whole MIMIC-III corpu, while AKI-BC-BERT is obtained by further pre-training BC-BERT on AKI corpus which is a subset of MIMIC-III corpus. Comparing the results of BC-BERT and AKI-BC-BERT, AKI-BC-BERT performs better than BC-BERT in most cases, which indicates that further pre-training on a subset of the clinical notes can still improve the performance for the subset-related tasks, i.e., AKI early prediction in our consideration. AKI-BioBERT is obtained by pre-training BioBERT on the AKI-related subset of MIMIC-III corpus. The difference between AKI-BioBERT and BC-BERT is the pre-training dataset. Comparing the results of BC-BERT and AKI-BioBERT, AKI-BioBERT also performs better than BC-BERT in most cases, that is because pre-training on the whole MIMIC-III corpus does not perform as good as on its AKI-related subset. This may suggest that the MIMIC-III corpus without the AKI-related subset cannot help the AKI early prediction task, and even harmful, which is also validated by the comparison of results of Clinical BERT and AKI-baseBERT that AKI-baseBERT performs better than or comparable to Clinical. Also, from Table \ref{tab:results}, the results of AKI-BioBERT are better than AKI-baseBERT in most cases, which indicates that the biomedical corpora sourced from PubMed can indeed improve the model for AKI early prediction. Among all the 7 models considered, only the pre-training processes of BioBERT and BERT-base have not involved the AKI-related corpus, thus the results of the two models are worse than the other five models as expected.

{Comparing different balancing strategies of AKI-BERT in Table \ref{tab:results}, in most cases of pooling, SBS performs better than DS, US and weight in terms of AUC, but weight performs best in terms of F1. Although the weight balancing strategy performs good for BERT models that are further pre-trained on AKI-related corpus, it fails for other BERT models like BC-BERT, Clinical BERT, BioBERT and BERT-base. And SBS strategy also fails for BioBERT and BERT-base. Given that weight$>$SBS$>$DS=US in terms of imbalance degree, this may suggest that imbalanced batches rely more on good initialization of BERT even if a class weight balancing strategy is used. }

As we can see AKI-BERT can outperform the corresponding base BERT models consistently for AKI early prediction, thus, the pre-trained model can be helpful for practitioners that focus on the prediction performance. On the other hand, although the pre-training process consumes much time (3 days or so with Tesla V100 GPU), we release our pre-trained models that are ready for fine-tuning for the downstream AKI-related tasks without pre-training again. We believe that releasing our pre-trained models will be useful to the community.

\subsection{Attention Visualization}
\begin{figure}[!t]
\subfloat[AKI-BC-BERT]{\includegraphics[width=\linewidth]{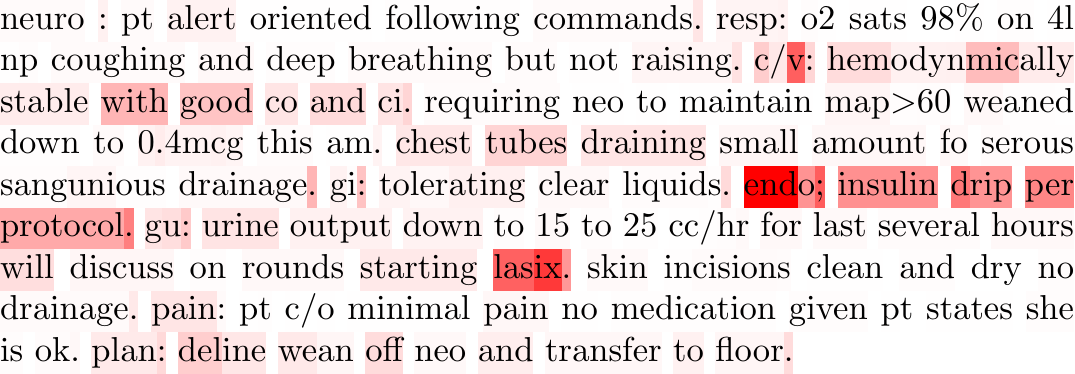} \label{fig:attakibcbert}}

\subfloat[BC-BERT]{\includegraphics[width=\linewidth]{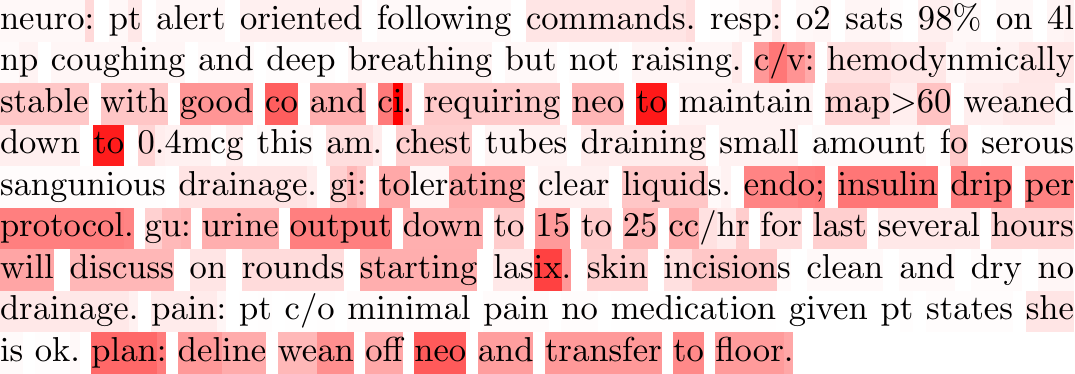} \label{fig:attbcbert}}

\caption{Attention visualization on an example note whose label is AKI.}
\label{fig:attention}
\end{figure}

We show the attention visualization of each token in an example notes in Figure \ref{fig:attention}. Figure \ref{fig:attakibcbert} and \ref{fig:attbcbert} correspond to the AKI-BC-BERT and BC-BERT, respectively. In Figure \ref{fig:attention}, the highlight indicates the importance of the words for the label, the darker the highlight, the more important the words are for the AKI early prediction task. {We can see that AKI-BC-BERT pays more attention to the words `lasix', `endo', `insulin drip', and `protocol' in this case, which is consistent with the previous finding using bag of words by Li et al. \cite{li2018early}; yet, BC-BERT spreads the attention over more words than AKI-BC-BERT; some of the words (e.g., `to' and `plan') are not related to AKI in a general sense.} Comparing the highlight annotations between AKI-BC-BERT and BC-BERT, the annotation of AKI-BC-BERT is more precise than BC-BERT, which demonstrates the efficacy of pre-training on AKI-related corpus.

\section{Conclusion}
In this paper, we explored the pre-trained contextual language model BERT on AKI domain notes. We presented AKI-BERT that was further pre-trained from a pre-trained BERT on AKI domain corpus. We found domain-specific AKI-BERT can achieve better performance than BERT-base, Clinical BERT or BioBERT for AKI early prediction. We also found the pooling strategy is more effective than simple truncating for long notes. The limitation is that we only use MIMIC-III ICU notes which is from a single healthcare institution. Language styles across different institution are usually significantly different, we need more AKI notes from various institution to pre-train an AKI-BERT model suitable for various institution. In the future work, we will explore AKI-BERT for other medical NLP tasks such as AKI progression and mortality prediction. Leveraging domain knowledge to guide the fine-tuning of BERT is another interesting direction.

%

\section*{Declarations}

\begin{backmatter}

\section*{Ethics approval and consent to participate}
Not applicable.
\section*{Consent to publish}
Not applicable.
\section*{Availability of data and materials}
We released our code and models at \url{https://github.com/mocherson/AKI_bert}. The datasets generated and/or analyzed during the current study are available in the MIMIC repository, \url{https://mimic.physionet.org/}.
\section*{Competing interests}
The authors declare that they have no competing interests.
\section*{Funding}
This work is supported by NIH Grant R01LM013337.
\section*{Authors' Contributions}
CM and YL originated the study and wrote the manuscript. LY and CM designed the method framework. LY provided the idea of pre-training and fine-tuning for BERT. CM conducted the experiments. YL and CM contributed to the results analysis and discussion. YL guided the study. All authors read and approved the manuscript.
\section*{Acknowledgements}
Not applicable.


\bibliographystyle{bmc-mathphys} 
\bibliography{aki-bert}      

\end{backmatter}
\end{document}